\pdfoutput=1
\documentclass[11pt]{article}

\usepackage[final]{acl}

\usepackage{times}
\usepackage{latexsym}
\usepackage{hyperref}

\usepackage[T1]{fontenc}
\usepackage[utf8]{inputenc}

\usepackage{microtype}

\usepackage{inconsolata}

\usepackage{graphicx}

\title{Towards Zero-Shot Text-To-Speech for Arabic Dialects}

\author{
  Khai Duy Doan$^{\xi}$\,
  ~~~~~Abdul Waheed$^{\xi}$\,
  ~~~~~Muhammad Abdul-Mageed$^{\xi,\gamma,\lambda}$ \\
  $^{\xi}$MBZUAI~~~~~ 
  $^{\gamma}$The University of British Columbia ~~~~~$^\lambda$ Invertible AI \\
  \texttt{\normalsize \{duy.doan,abdul.waheed\}@mbzuai.ac.ae}~~~~~~~~\texttt{\normalsize muhammad.mageed@ubc.ca}
}

\begin{document}
\maketitle

\begin{abstract}
Zero-shot multi-speaker text-to-speech (ZS-TTS) systems have advanced for English, however, it still lags behind due to insufficient resources. We address this gap for Arabic, a language of more than 450 million native speakers, by first adapting a sizeable existing dataset to suit the needs of speech synthesis. Additionally, we employ a set of Arabic dialect identification models to explore the impact of pre-defined dialect labels on improving the ZS-TTS model in a multi-dialect setting. Subsequently, we fine-tune the XTTS\footnote{https://docs.coqui.ai/en/latest/models/xtts.html}\footnote{https://medium.com/machine-learns/xtts-v2-new-version-of-the-open-source-text-to-speech-model-af73914db81f}\footnote{https://medium.com/@erogol/xtts-v1-techincal-notes-eb83ff05bdc} model, an open-source architecture. We then evaluate our models on a dataset comprising 31 unseen speakers and an in-house dialectal dataset. Our automated and human evaluation results show convincing performance while capable of generating dialectal speech. Our study highlights significant potential for improvements in this emerging area of research in Arabic.
\end{abstract}
\section{Introduction}\label{sec:introduction}

Over the last few years, there have been dramatic breakthroughs in speech synthesis. The core of any text-to-speech (TTS) system lies in its capacity to transform textual input into natural, coherent speech. This transformation, however, is not a simple task. It involves the integration of multiple components. Understanding and dissecting these components - from text preprocessing and linguistic analysis to acoustic modeling and prosody generation - is essential for enhancing the quality and naturalness of synthesized speech and harnessing the full potential of TTS technology in diverse applications.

While the latest TTS systems, \cite{hifigan}, \cite{gradtts}, \cite{kim2020glowtts}, \cite{diffwave}, can synthesize high-quality speech from single or multiple speakers, it is still a challenge to synthesize speech from unseen speakers. Transiting from single-speaker to multi-speaker TTS, the idea of encoding the speaker information with low dimensional embeddings is applied. However, in the multi-speaker setting, models can only generate the speech of the speakers including in the training dataset. Subsequently, generating the voices of an unseen speaker based on a few seconds of the speaker's utterance has emerged as a challenging task. This approach is often referred  as \textit{zero-shot multi-speaker TTS (ZS-TTS)}, defined as a task that aims to synthesize voices for new speakers not encountered during training, employing only a few seconds of speech.

\begin{figure}
    \centering
    \includegraphics[scale=0.4]{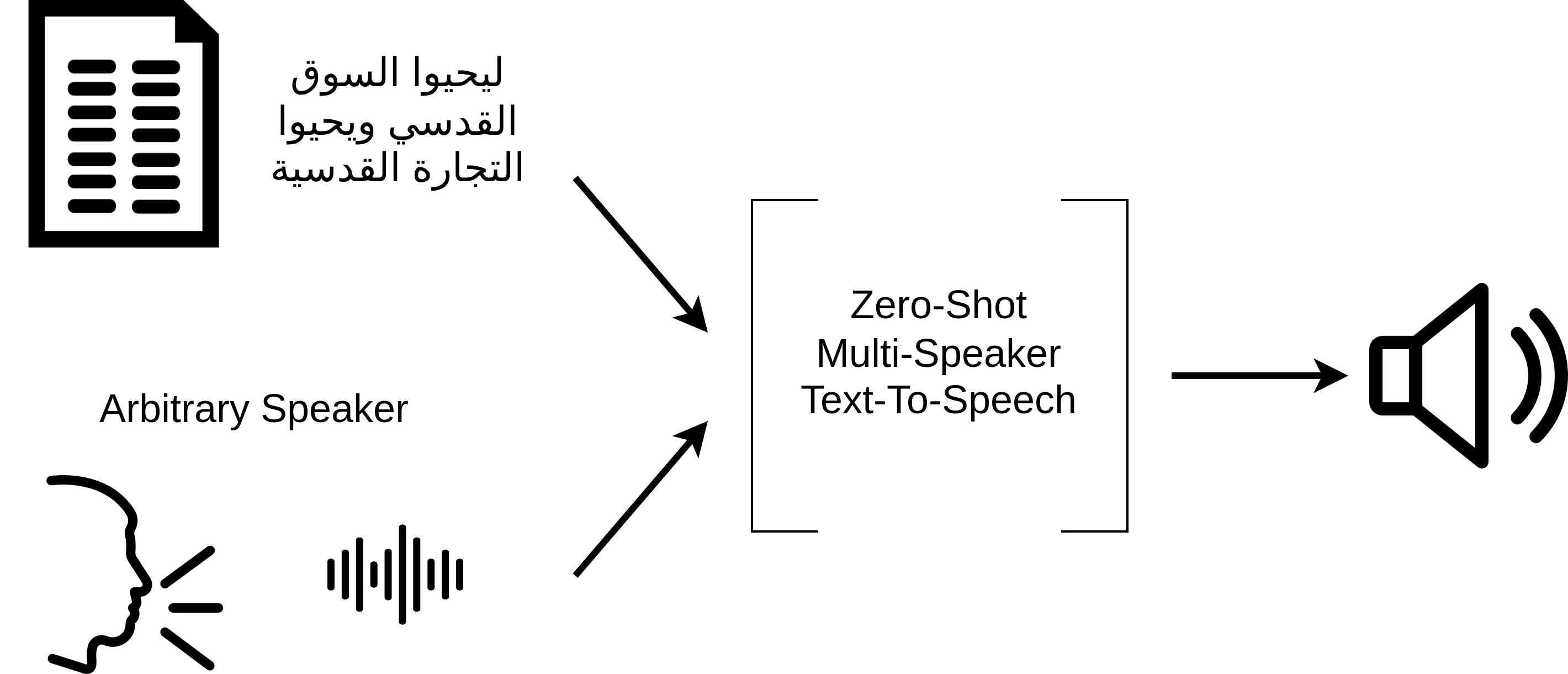}
    \caption{A zero-shot multi-speaker TTS system. It synthesizes speech based on the input text as well as preserves the acoustic features of input speech segment.}
    \label{fig:zs-tts}
\end{figure}

On the other hand, the research landscape for TTS in Arabic remains relatively unexplored, particularly ZS-TTS systems. ASC \cite{asc} is considered the first open-source Arabic dataset curated for speech synthesis, while~\cite{clartts} presents a speech corpus for Classical Arabic TTS (ClArTTS). In \cite{ALRADHI2020101025}, the authors proposed a continuous residual-based vocoder for speech synthesis using a small audio-visual phonetically annotated Arabic corpus. Recently, TunArTTS \cite{laouirine-etal-2024-tunartts-tunisian} addresses the lack of development of end-to-end TTS systems for the Tunisian dialect by presenting a small dataset including 3+ hours of a male speaker. Furthermore, \cite{artst} proposes ArTST, a pre-trained Arabic text and speech transformer, followed by SpeechT5 \cite{speecht5}. ArTST utilizes ClArTTS for TTS fine-tuning. Additionally, XTTS\footnote{https://docs.coqui.ai/en/latest/models/xtts.html}, presented by CoquiAI\footnote{https://github.com/coqui-ai}, has the voice cloning ability in different languages, including Arabic. Nevertheless, the dataset used to train XTTS is not revealed. Although some works are carried out for Arabic TTS, there has been no public research for Arabic ZS-TTS as of the time we are conducting this research, particularly in multi-dialect settings. In this paper, we offer the following contributions:
\begin{itemize}
    \item We re-target an available corpus, QASR~\cite{qasr} to suit the needs of ZS-TTS in Arabic, acquiring a rich resource for this task. 
    \item We develop highly effective ZS-TTS models for Arabic through fine-tuning XTTS on the refined QASR corpus, incorporating supplementary Arabic dialect labels. 
    % \item We make the dataset and our models publicly available.
\end{itemize}

\section{Related Works}\label{sec:related_works}

% Few-shot Voice Cloning

Sercan et al. \cite{fewshotvoicecloning} pioneered work on transmitting from multi-speaker to few-shot TTS. They dive into the task of voice cloning within sequence-to-sequence neural speech synthesis systems that learn to synthesize a person's voice from a few audio samples. 
% Zero-shot multi-speaker text-to-speech with state-of-the-art neural speaker embeddings
Other work such as ~\cite{zeroshot} applies transfer learning for ZS-TTS by extending an improved Tacotron system to a multi-speaker TTS framework and analyzing the effectiveness of different speaker embeddings in capturing and modeling characteristics of unseen speakers.
% Attentron: Few-Shot Text-to-Speech Utilizing Attention-Based Variable-Length Embedding
The framework of~\cite{zeroshot} has a challenge to produce a single embedding to present every utterance characteristic. Attentron~\cite{attentron}, a novel architecture designed for few-shot text-to-speech systems, was proposed to overcome this challenge.
% Normalization Driven Zero-Shot Multi-Speaker Speech Synthesis
In addition,~\cite{normalizationdrivenzeroshot} was introduced to better capture the prosody of unseen reference speech such as speaker embedding, pitch, and energy in a zero-shot manner.

% Meta-StyleSpeech
% Although Meta-StyleSpeech \cite{metastylespeech} has addressed some of the problems of previous

Although the performance of previous works is adequate to some extent, they still suffer from some problems. For example, in the pre-training and fine-tuning framework, the model is pre-trained on a large dataset and then fine-tuned with a few audio samples of the target speaker, which is not applicable to real-world setting. On the other hand, for those methods that are trained conditionally on a latent style vector they heavily rely on the diversity of the speakers in the source dataset. Meta-StyleSpeech~\cite{metastylespeech} addressed these problems by incorporating Style-Adaptive Layer Normalization (SALN) with two discriminators trained with style prototypes and episodic training techniques.
% SC-GlowTTS: an Efficient Zero-Shot Multi-Speaker Text-To-Speech Model
Following the development trend of TTS systems, SC-GlowTTS \cite{scglowtts} explores a flow-based decoder that works in a zero-shot scenario. Combining GlowTTS \cite{kim2020glowtts} and HiFi-GAN \cite{hifigan}, SC-GlowTTS is capable of synthesizing speech faster than in real-time.
% Grad-StyleSpeech
Adapting the advancement of score-based diffusion model \cite{SDE}, \cite{gradstyle} present Grad-StyleSpeech. This model excels in generating highly natural speech with an exceptionally high resemblance to the voice of a target speaker. 
% Guided-TTS 2
Moreover, applying a score-based diffusion model, Guided-TTS 2 \cite{guidedtts2} can be successfully fine-tuned with data of target speakers without transcription.

% YourTTS
The aforementioned works, however, have not yet explored the multilingual TTS, which can be useful to ZS-TTS in that it can diversify training data. One exception is YourTTS \cite{yourtts}, which emerged as the first work proposing a multilingual approach to zero-shot multi-speaker TTS.
% VALLE
Other works include VALL-E~\cite{valle}, VALL-E X~\cite{vallex}. Instead of mel-spectrogram, VALL-E \cite{valle} utilizes discrete audio codec code \cite{encodec} as speaker representation. VALL-E is trained with LibriLight, which is a large and diverse dataset including 60k hours of English speech with over 7000 unique speakers.
% VALLE-X
Soon afterward, the authors published VALL-E X~\cite{vallex} for cross-lingual speech synthesis. The multi-lingual in-context learning framework empowers VALL-E X to generate cross-lingual speech while preserving the characteristics of the unseen speaker, including their voice, emotion, and speech background, using only a single sentence in the source language.

% Voicebox
Inspired by large-scale generative models such as GPT and DALLE, Voicebox \cite{voicebox} was also presented. Voicebox is trained on a text-guided speech-infilling task by generating masked speech given surrounding audio and a text transcript. Voicebox's text-guided infilling method demonstrates superior scalability in terms of data, covering various speech-generative tasks.

\section{XTTS Model}\label{sec:methodology}
XTTS~\cite{Eren_Coqui_TTS_2021}~\footnote{We refer to XTTSv2 simply as XTTS in our study.} is a multilingual multispeaker speech synthesis model that supports Arabic (ar) along with English (en), and multiple other languages.

Overall, XTTS can generate natural-sounding audio with as short as three seconds of reference speech across 15 languages. It is based on autoregressive modeling trained on discrete audio representations. The model has a VQ-VAE to discretize audio into discrete tokens and subsequently uses GPT2~\cite{gpt2} model to predict these audio tokens based on the input text conditioned upon latent speaker representation. The computation of speaker latent is facilitated by the Perceiver model~\cite{jaegle2021perceiver}, which takes mel spectrogram as inputs and produces latent vectors representing speaker information to prefix the GPT decoder. XTTS uses the HifiGAN model to compute the final audio signal from the GPT2 outputs which reduces the inference latency.

The empirical findings indicate that the Perceiver model outperforms simpler encoders, such as Tortoise~\cite{betker2023better}, and speech prompting models like Vall-E~\cite{wang2023neural}, in capturing speaker characteristics. Notably, the Perceiver exhibits a remarkable consistency in model outputs across various runs, mitigating the occurrence of speaker shifting between different model executions. In addition to that, the intrinsic flexibility of the Perceiver is highlighted by its ability to accommodate multiple references without imposing any length constraints. This feature facilitates the comprehensive capture of diverse facets of the target speaker, and the potential to amalgamate characteristics from different speakers, thereby enabling the creation of a distinctive and unique voice. In XTTS, input text undergoes BPE tokenization, and a language token is inserted at the beginning, resulting in a structured sequence: \textit{[bots], [lang], t1, t2, ..., tn, [eots]}. Speaker latents are obtained from a mel-spectrogram and these vectors directly condition the model. Training samples are formed by appending audio tokens to the input sequence: \textit{s1, s2, ..., sk, [bots], [lang], t1, t2, ..., tn, [eots], [boas], a1, a2, ..., a\_l, [eoas]}. The three-stage training process for XTTS involves VQVAE, GPT, and audio decoder training. The three components of XTTS model are demonstrated in Appendix, Fig.\ref{fig:xtts}.

\begin{figure}[]
  \includegraphics[width=\columnwidth]{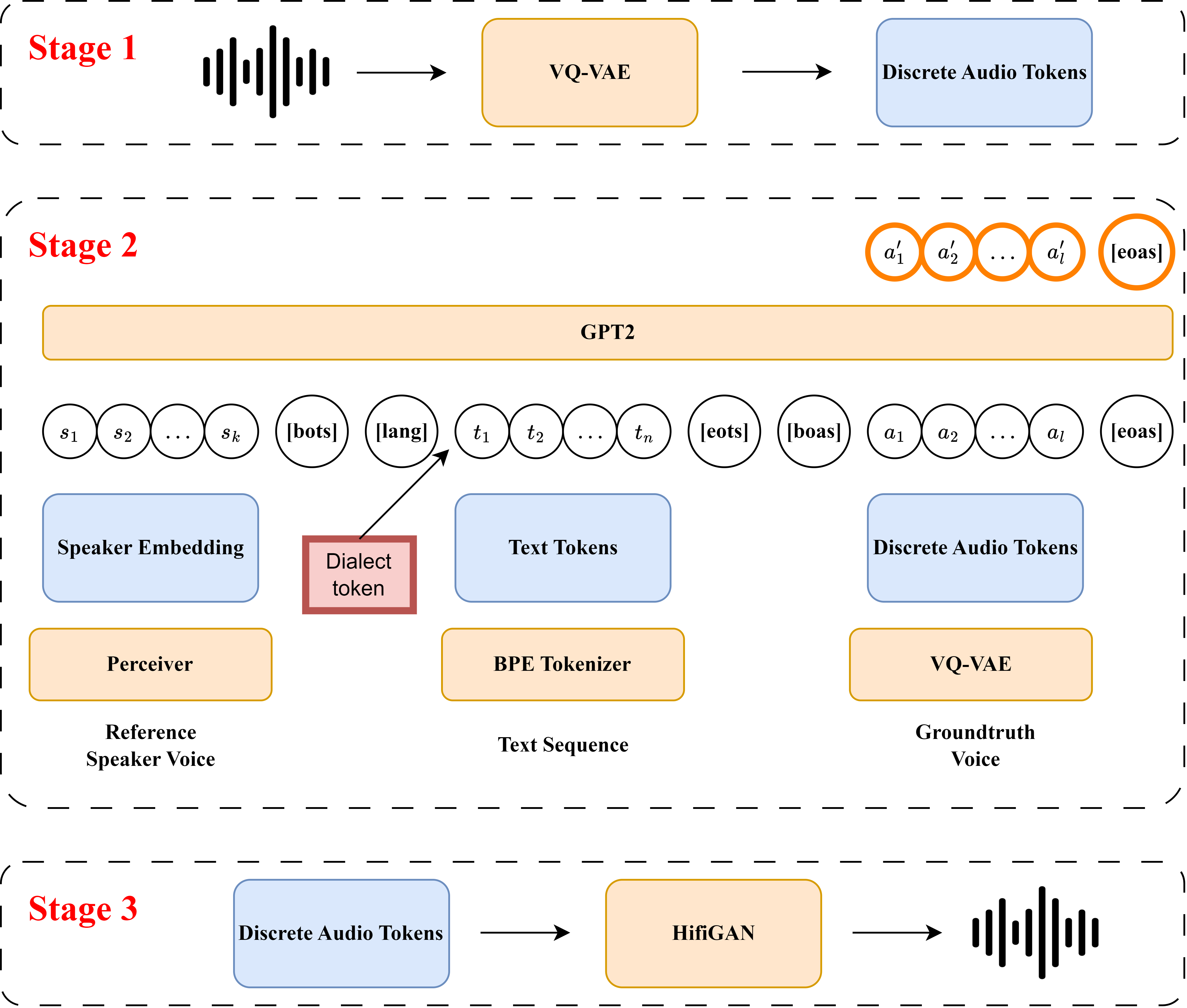}
  \caption {The three stages of training XTTS model. \textbf{Stage 1.} Train the VQ-VAE to learn the discrete audio tokens representation from audio waveform. \textbf{Stage 2.} A GPT2-based model is trained to autoregressively generate discrete audio tokens from the concatenation of speaker, text, and discrete audio tokens embeddings. \textbf{Stage 3.} HifiGAN is trained as a decoder to generate audio wavefrom from the discrete audio tokens, generated by the GPT2.}
  \label{fig:xtts}
\end{figure}

XTTS is trained on publicly available datasets across different languages. Since publicly available data varies for different languages, we believe that Arabic has a relatively small amount of data compared to other high-resource languages such as English, German, and Spanish that it supports. Although the outputs from XTTS seem high quality, developers of the model do not conduct any evaluation for any languages. To address this gap, we first train XTTS on approximately ~2000 hours of high-quality Arabic speech described in~\ref{sec:dataset}. We then perform a comprehensive evaluation using both an automatic metric as well human evaluation.

\section{Experiments}\label{sec:experiments}

\subsection{Dataset}\label{sec:dataset}

We use QASR \cite{qasr} as the training data which contains approximately 2,000 hours of speech sampled at 16kHz crawled from the Aljazeera news channel. QASR involves multiple dialects, no label yet, and includes 1.6 million segments of 27,977 speakers. Transcripts in QASR are aligned with audio segments using light supervision. One of the primary considerations for selecting QASR for our experiments is that it has the speaker identity of audio recordings.

\begin{figure}[h]
    \centering
    \includegraphics[width=0.5\textwidth]{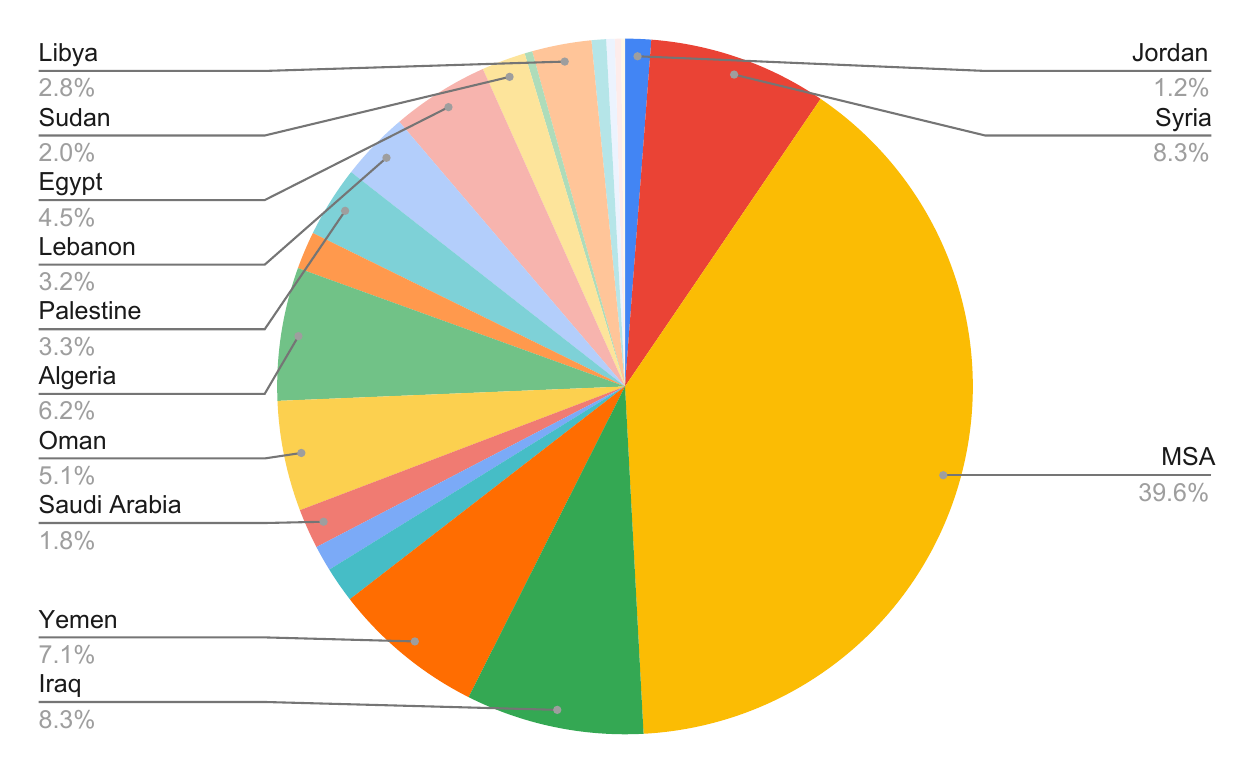}
    \caption{\label{fig:dialect-distribution}
    QASR data distribution across different dialects.
    }
\end{figure}

\noindent\textbf{Text-Audio Processing.}~Because the audios of the QASR dataset is originally crawled from a news channel, we need to remove the background noise to make the dataset suitable for training the TTS system. We employ the FRCRN \cite{FRCRN} system\footnote{\url{https://github.com/alibabasglab/FRCRN}} to suppress the background noise of speech audios. Moreover, when exploring the dataset, we found that some text-audio pairs are mismatched. The problem occurs when two people are speaking at the same time in the audio segment. To filter out these mismatched samples, we first employ the Whisper~\cite{whisper} model to generate the transcripts of all audio segments. Subsequently, we compute the word error rate (WER) between the generated and ground-truth transcripts. We then remove all samples with WER larger than zero.

\noindent\textbf{Dialect Pseudo Labelling.}
We train our multi-dialect XTTS model on QASR, however, the original QASR data does not have a dialect associated with each segment. To overcome this we employ eight dialect identification models to predict the dialect from the text then we take the label with the highest cumulative confidence as the final dialect label. The models are collected from Huggingface \footnote{\url{https://huggingface.co/AMR-KELEG/ADI-NADI-2023}}\footnote{\url{https://github.com/Lafifi-24/arabic-dialect-identification}}\footnote{\url{https://huggingface.co/CAMeL-Lab/bert-base-arabic-camelbert-mix-did-madar-corpus26}}\footnote{\url{https://huggingface.co/CAMeL-Lab/bert-base-arabic-camelbert-msa-did-madar-twitter5}}. We provide data distribution across dialects in Figure~\ref{fig:dialect-distribution}

\noindent\textbf{Train-Test Splits.}\label{train-test-split}~After the processing steps, there are 17341 remaining speakers, of which we use 17310 speakers for training and 31 speakers for evaluation. Those 31 speakers are excluded from training for zero-shot setting evaluation.

% \noindent\textbf{Train-Test Splits.}\label{train-test-split}~ After the processing steps, there are 27,590 remaining speakers, of which we use 27,550 speakers for training and 40 speakers for evaluation. Those 40 speakers are excluded from training for zero-shot setting evaluation.

In addition to testing the split of QASR, we use a small amount of in-house dialect data to evaluate out-of-distribution generation quality. We collect this data from YouTube and transcribe them using native speakers from the respective dialects. Our in-house dialectal data includes country-level dialects from Algeria, Jordan, Palestine, UAE, and Yemen.

\subsection{Methods}

In addition to the pre-trained XTTS model, we develop two models by fine-tuning XTTS on the curated QASR dataset:

\noindent\textbf{Fine-tuned without dialect token.} We fine-tune the model with the same parameters, vocabulary as the pre-trained model.

\noindent\textbf{Fine-tuned with dialect token.} We add a list of 22 dialect tokens (21 dialects and MSA) to the vocabulary so the vocabulary size increases from 6681 to 6703 entries. To provide the model with the dialect information, we insert the dialect token after the \textit{[lang]} token: \textit{s1, s2, ..., sk, [bots], [lang], \textbf{[dialect]} t1, t2, ..., tn, [eots], [boas], a1, a2, ..., a\_l, [eoas]}. In the embedding layer, we initialize the weights for new tokens by sampling from the standard normal distribution.

% \begin{itemize}
%     \item 
%     \item \textit{}: 
% \end{itemize}

We train the models for 300,000 steps on two NVIDIA A100-SXM4 40GB with a batch size of 6 and gradient accumulation steps of 38. We run the evaluation with three settings: the pre-trained XTTS model as \textit{baseline}, \textit{fine-tuned without dialect token}, and \textit{fine-tuned with dialect token}.

\subsection{Evaluation}

\textbf{Speaker Similarity.} We use speaker embedding cosine similarity  (SECS) as the quantitative metric evaluation. To compute the speaker embedding vector of speech audio, we employ the \href{https://github.com/resemble-ai/Resemblyzer}{Resemblyzer} package. Afterwards, we get the cosine similarity score between two speaker embedding vectors, which is in the range [-1, 1], where a larger value indicates a higher similarity of input vectors. For each sample synthesis, we randomly pick a reference utterance, which has at least 5-second duration length, from the same speaker. We run the experiment \textit{three times and report the average} score.

\noindent\textbf{Word Error Rate.} In order to quantify the \textit{consistency} of generated voice with the ground-truth transcript, we employ the Word Error Rate (WER). We extract the transcript from generated audio by using Whisper-large~\cite{whisper}. The package Jiwer~\footnote{\url{https://github.com/jitsi/jiwer}} is used to compute the WER.

\noindent\textbf{Human Evaluation.} We also conduct human evaluation to quantify the quality of our models when it comes to \textit{naturalness} and \textit{fluency}. To conduct the human evaluation, we first prepare a guideline that explains the task to human annotators, including the ranking system employed (between 1 and 5, with increments of 0.5) where one is lowest and 5 is highest. We then propagate the synthesized audio for annotators to start the job. We generate ten audio segments from each model and employ five speakers for evaluation. We conduct the human evaluation only on the in-house dialect dataset, which was labeled under strict supervision. Check Appendix~\ref{sec:appendix} to listen to the generated audio samples.
\section{Results and Discussion}

% \textcolor{red}{Update MOS score}

\begin{table}
  \centering
  \begin{tabular}{lcc}
    \hline
    \textbf{Model} & \textbf{WER} & \textbf{SECS} \\
    \hline
    baseline  & \textbf{6.42} & 0.755 \\ \hline
    ft w/o dialect token   & 16.79 & \textit{0.766} \\
    ft w/ dialect token    & 17.96 & \textbf{0.766} \\\hline
  \end{tabular}
  \caption{Evaluation results on QASR unseen speakers.}
  \label{tab:qasr_eval}
\end{table}

\begin{table}
  \centering
  \begin{tabular}{lccc}
    \hline
    \textbf{Model} & \textbf{WER} & \textbf{SECS} & \textbf{MOS} \\
    \hline
    baseline  & \textbf{47.16} & 0.79 & \textbf{3.61} \\ \hline
    ft w/o dialect token   & \textit{57.47} & \textit{0.806} & \textit{3.53} \\
    ft w/ dialect token    & 62.74 & \textbf{0.825} & 3.19 \\\hline
  \end{tabular}
  \caption{Evaluation results on Speech-Home-Dataset with real data of Arabic dialects.}
  \label{tab:dialect_eval}
\end{table}

We fine-tune the XTTS model with two settings on roughly 472 hours of multi-dialectal Arabic speech data. We report automatic evaluation metrics SECS and WER on held-out data~\ref{train-test-split} in Table~\ref{tab:qasr_eval}. Our automatic evaluation reveals that while \textit{baseline} outperforms the other two models in terms of WER (6.42 vs 16.79 and 17.96), the SECS shows that the two fine-tuned models are superior. The same scenario happens on the in-house Arabic dialect dataset, as shown in Table~\ref{tab:dialect_eval}. This result shows that while the fine-tuned models are better in preserving the acoustic features of the reference speaker, they mess up in mapping the text units to the corresponding speech units. Although we have filtered out samples with high WER based on the Whisper model, we believe that the training dataset requires further refinement to improve the performance of the models.

In addition to the held-out dataset, we probe the performance of all models on the clean, labeled with strict supervision, in-house dialect dataset. We report the WER, SECS, and MOS score in Table~\ref{tab:dialect_eval}. Based on our automatic evaluation, we find that with the dialect dataset, the setting \textit{fine-tuned with dialect token} outperforms the other models with a larger margin (0.035) than that of QASR held-out samples (0.011). In addition to this, we also conducted human evaluation on dialect data and we found that fine-tuning with dialect gives on-par results compared to the supervised baseline, while also being capable of generating dialectal speech. This suggests that even though we only use the pseudo dialect label, the dialect information does help to improve the quality of generated speech for the Arabic dialect. % Further, we note that when removing diacritics from input text the original XTTS models improve while others get worse. This suggests that the original XTTS might have been trained in modern standard Arabic (MSA) where diacritics are often ignored.
\section{Conlusion}

Generating speech from text has many critical applications particularly for individuals with visual and hearing impairments, enabling them to access and interact with digital content through screen readers, spreading across different regions and languages. However, previous research focuses on building text-to-speech systems for high-resource languages and often lacks comprehensive evaluation. In this work, we address this gap by fine-tuning two models for Arabic text-to-speech generation. We then conduct a thorough evaluation and comparative analysis of pre-trained with ours. Findings reveal that although our models are inconsistent in mapping phonemes to sounds, they more precisely preserve acoustic features compared to the baseline.

In future work, we intend to train our models on extensive high-quality data while ensuring that the training data is representative of the diverse linguistic varieties of Arabic. For instance, we would like to extract a subset of the QASR dataset and perform manual labeling to create a small, clean dataset. Subsequently, we will employ semi-supervised learning techniques to enhance our models' performance. Finally, upon acceptance of our work, we intend to release the model weights, inference pipeline, and training protocols.

% incorporate dialect-specific information during training so that text-to-speech models can handle various Arabic dialects. We intend to release the model weights, inference pipeline, and training recipes upon acceptance of our work. 

\section{Limitations}

Despite the models are capable of generating natural-sounding speech, comprehensible by native speakers, which is often indistinguishable from reference speakers, our work has some clear limitations which we discuss below:
\begin{itemize}
    \item \textbf{Data.} Gender-wise distribution is severely imbalanced. For example, 69\% of 1.6M segments are male while only 6\% are female and the rest remain unknown. This also limits our model's capability to generate high-quality speech for female reference speakers. 
    \item \textbf{Pseudo Dialect Label.} Although nine dialect identification models were used to assign dialect labels to each sample, these labels are considered pseudo-labels. We believe that obtaining more accurate labels will improve the model's performance.
    \item \textbf{Training.} While we train our models on 300,000 steps with a cumulative batch size of 228, the original models are trained for significantly more steps (data). To overcome this, we continuously train our models and intend to update the results as we get better checkpoints.
\end{itemize}

\section*{Acknowledgments}\label{sec:acknow}
We acknowledge support from Canada Research Chairs (CRC), the Natural Sciences and Engineering Research Council of Canada (NSERC; RGPIN-2018-04267), the Social Sciences and Humanities Research Council of Canada (SSHRC; 895-2020-1004; 895-2021-1008), Canadian Foundation for Innovation (CFI; 37771), Digital Research Alliance of Canada,\footnote{\href{https://alliancecan.ca}{https://alliancecan.ca}} and UBC Advanced Research Computing-Sockeye.\footnote{\href{https://arc.ubc.ca/ubc-arc-sockeye}{https://arc.ubc.ca/ubc-arc-sockeye}}

\bibliography{custom}

\appendix

\section{Appendix}\label{sec:appendix}

\textbf{XTTS baseline}

\url{https://arabicnlp.aidaform.com/baseline}

\noindent\textbf{Fine-tuned without dialect token}

\url{https://arabicnlp.aidaform.com/fine-tuned-without-dialect-token}

\noindent\textbf{Fine-tuned with dialect token}

\url{https://arabicnlp.aidaform.com/fine-tuned-with-dialect-token}

% \begin{figure*}[t]
%   \includegraphics[width=0.48\linewidth]{example-image-a} \hfill
%   \includegraphics[width=0.48\linewidth]{example-image-b}
%   \caption {A minimal working example to demonstrate how to place
%     two images side-by-side.}
% \end{figure*}

\end{document}